# OdoNet: Untethered Speed Aiding for Vehicle Navigation Without Hardware Wheeled Odometer

Hailiang Tang, Xiaoji Niu, Tisheng Zhang, You Li and Jingnan Liu

*Abstract*—Odometer has been proven to significantly improve the accuracy of the Global Navigation Satellite System / Inertial Navigation System (GNSS/INS) integrated vehicle navigation in GNSS-challenged environments. However, the odometer is inaccessible in many applications, especially for aftermarket devices. To apply forward speed aiding without hardware wheeled odometer, we propose OdoNet, an untethered one-dimensional Convolution Neural Network (CNN)-based pseudo-odometer model learning from a single Inertial Measurement Unit (IMU), which can act as an alternative to the wheeled odometer. Dedicated experiments have been conducted to verify the feasibility and robustness of the OdoNet. The results indicate that the IMU individuality, the vehicle loads, and the road conditions have little impact on the robustness and precision of the OdoNet, while the IMU biases and the mounting angles may notably ruin the OdoNet. Thus, a data-cleaning procedure is added to effectively mitigate the impacts of the IMU biases and the mounting angles. Compared to the process using only non-holonomic constraint (NHC), after employing the pseudo-odometer, the positioning error is reduced by around 68%, while the percentage is around 74% for the hardware wheeled odometer. In conclusion, the proposed OdoNet can be employed as an untethered pseudo-odometer for vehicle navigation, which can efficiently improve the accuracy and reliability of the positioning in GNSS-denied environments.

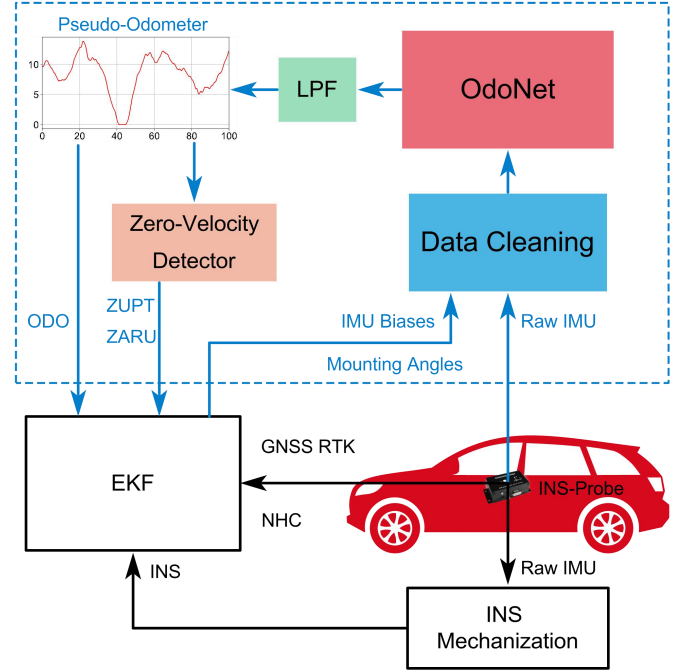

Fig. 1. An overview of the proposed vehicle integrated navigation system based on the OdoNet.

## I. INTRODUCTION

THE GNSS/INS (Global Navigation Satellite System, Inertial Navigation System) integrated navigation system can provide full navigation parameters, including position, velocity, and attitude, and thus has been widely used in land vehicles. With the wide establishment of the ground-based augmentation system, we can obtain high-accuracy GNSS RTK (Real Time Kinematic) positioning in open-sky areas. However, GNSS can be easily interfered with or interrupted in urban environments. Inertial measurement units (IMU) can work independently, and INS can provide continuous navigation service during GNSS outages, which greatly improves the usability of the integrated navigation system. Due to lower cost and lower power consumption, low-grade MEMS (Micro-Electro-Mechanical System) IMU has been widely applied to vehicle navigation. Being restricted by the poor precision and the low stability of the MEMS IMU, the integrated navigation system cannot maintain high-accuracy positioning in GNSS-denied environments, especially when the vehicle travels through city canyons, long tunnels, and underground roads, in which GNSS tend to have long outages. Consequently, extra aiding sources or constraints, such as zero velocity update (ZUPT) [1], non-holonomic constraint (NHC) [2-4], wheeled odometer [4], and *et. al*, must be employed. ZUPT and NHC are kinds of virtual velocity observations from the vehicle or the IMU itself, which can be achieved independently. However, to use odometer aiding, we must attach an extra encoder to the wheel, or grab speed message

This research is funded by the National Key Research and Development Program of China (No. 2020YFB0505803), and the National Natural Science Foundation of China (No. 41974024). *(Corresponding author: Tisheng Zhang)*
Hailiang Tang is with the GNSS Research Center, Wuhan University, Wuhan 430079, China.
Xiaoji Niu, Tisheng Zhang, and Jingnan Liu, are with the GNSS Research Center, Wuhan University, Wuhan 430079, China, and also with the Collaborative Innovation Center of Geospatial Technology, Wuhan University, Wuhan 430079, China (e-mail: zts@whu.edu.cn).
You Li is with the State Key Laboratory of Information Engineering in Surveying, Mapping and Remote Sensing, Wuhan University, Wuhan 430079, China.

from the CAN (Controller Area Network) inside the vehicle, which greatly increases the cost and complexity of the system integration. As a consequence, the wheeled odometer cannot be applied in many applications, especially for aftermarket devices, which leads to poor positioning accuracy during GNSS outages, and thus significantly reduces the reliability of the integrated system in GNSS-denied environments.

Sufficient experimental explorations have demonstrated that it is possible to estimate the vehicle speed by learning from a single IMU. In this paper, we propose OdoNet, an untethered CNN (convolution neural network) based pseudo-odometer model, to directly estimate the vehicle speed. The framework of the proposed integrated system based on the OdoNet is depicted in Fig. 1. The main contributions of this paper are listed as follows:

● To apply forward speed aiding without hardware wheeled odometer, we propose a CNN-based pseudo-odometer to estimate the forward speed of the vehicle. The pseudo-odometer combined with an effective zero-velocity detector is employed in the integrated navigation system to improve the positioning accuracy in GNSS-denied environments.

● Since OdoNet is a data-driven method, its performance is strongly influenced by data quality. To fully verify the generalization capability and the precision of the OdoNet, dedicated experiments were carried out. The experiment results indicate that the IMU individuality, the vehicle loads, and the road conditions have little effect on the OdoNet, while the IMU biases and the mounting angles may notably ruin the OdoNet.

● To mitigate the impacts of the IMU biases and the mounting angles, a data-cleaning procedure is added to precisely compensated for these factors. The data-cleaning procedure is proven to significantly improve the robustness and the precision of the OdoNet.

The remainder of this paper is organized as follows: In the next section, a brief literature review on IMU-based deep learning network is presented. Then, the details of the OdoNet architecture and the integrated navigation system enhanced by it are described. The experiments and the results are presented to evaluate the precision and robustness of the proposed method. Finally, we conclude the proposed method.

## II. Related Works

The conventional GNSS/INS integrated navigation has been studied for decades, so only deep neural networks (DNN) will be discussed here. With the rapid development of DNN, IMU-based networks have gained intensive attention in positioning and navigation fields. According to the application of the deep learning network, we can classify these networks into two categories, including IMU-based odometry and velocity-estimation networks.

### A. IMU-based Odometry

Recently, more and more researchers have focused on the end-to-end DNN based IMU, such as inertial odometry [14-18] and visual-inertial odometry (VIO) [19-22]. The inertial odometry only uses IMU data to estimate the 6-DOF pose, while VIO uses both IMU and camera data.

For inertial-only odometry, Chen proposed IONet [14], a two-layer Bi-LSTM network to learn location transforms in polar coordinates from raw IMU data, but the network was only for 2D navigation. As future work, Chen released OxIOD [15], the dataset for deep inertial odometry. However, the dataset is only available for low-speed PDR applications. Yan [16] proposed three novel neural inertial navigation architectures for PDR (Pedestrian Dead Reckoning) applications on mobile phones, which were proved to outperform previous methods. Esfahani presented AbolDeepIO [17], a triple channel LSTM (long-short time memory) based DNN for vehicles, which could extract features from accelerometer, gyroscope, and time interval. However, the proposed model did not consider the vehicular characteristics. Afterward, an IMU-based 6-DOF inertial odometry was proposed in [18], which combined CNN with Bi-LSTM (Bidirectional Long-short Time Memory), but the experiments showed large increased error along the vertical axis.

DNN-based VIO, including supervised learning, unsupervised learning, and self-supervised, has also been studied for many years. VINet [19] was the first end-to-end deep learning VIO network, which applied CNN to extract features from the images and LSTM to process IMU, and used a core LSTM to combine visual and inertial data. VIOLearner [20-21] was an unsupervised deep network of VIO, which could correct errors online. In [22], a novel self-supervised deep learning VIO network was proposed, which adopted 3D geometric constraints and updated additional bias for IMU using the pose feedback. Many researchers have also focused on the deep learning network combining LiDAR and IMU [23-24].

DNN-based odometry substitutes neural networks for precise mathematical models, which completely changes the essence of navigation algorithms. For training a general DNN model, we must carry out heavy data collection work, and even so, we still cannot guarantee that the model can work every time in every scene. In practical application, there is still a lot of work to do for DNN-based odometry.

### B. Velocity Estimation

Using IMU data to estimate velocity has also become a research hotspot. Some of the researchers only construct a classification problem to detect some special states, like zero velocity and zero angular velocity states. Others employ IMU-based DNN to directly estimate velocity.

ZUPT is an effective approach to constrain INS drifts in PDR [1][5-7] and VDR [9] (Vehicle Dead Reckoning) applications. Xinguo *et al*. proposed AZUPT [5], a CNN-based zero-velocity detector, which could adaptively pick ZUPT points in different motion types. In [6-7], LSTM [8] based recurrent neural network (RNN) was used as a zero-velocity detector to improve the foot-mounted inertial navigation. Zero-velocity states can also be detected using regressed velocity from a DNN, and thus directly estimating velocity may be more effective and reasonable.

Brossard [9] applied an LSTM network to detect a variety of situations of interest on wheels, including zero velocity, zero angular velocity, zero lateral velocity, and zero vertical velocity. These constraints were treated as pseudo-measurements to refine the estimates of the IEKF (Invariant Extended Kalman Filter). However, if we model the lever-arms and the mounting angles of the IMU to the vehicle precisely, then the NHC (zero lateral velocity and zero vertical velocity) is typically satisfied in the non-steering wheel of the vehicle, except some special cases. In other words, DNN can be applied to estimate the vehicle speed or something else, not just NHC.

Moreover, the IMU-based DNN, which directly estimates the velocity, is also a research hotspot. A 2D CNN-based deep-learning model was used to predict 3D speed based on a window of IMU samples for mobile phone applications [10], and the regressed speed was used as a pseudo measurement in EKF (Extended Kalman Filter). RIDI [11] applied SVM (Support Vector Machine) and SVR (Support Vector Regression) to form a regression model and predicted velocity in smart-phone for correcting low-frequency biases in the accelerations, which were integrated twice to estimate position. A hybrid network [12-13] was proposed for PDR applications, which used a CNN to extract features from a signal stream, and a Bi-LSTM was used to estimate velocity from extracted features.

As one can see, most of the velocity estimating models are designed for PDR applications, which are not suitable for vehicle navigation. The vehicle has many characteristics that are completely different from mobile phones or foot-mounted devices. In addition, previous works show few pieces of evidence to prove the robustness and generalization capability of their network, which are the most important characteristics of a data-driven algorithm. For a vehicle integrated navigation system, what we need is an aiding source that can maintain high-accuracy positioning during GNSS signal blockage or degradation, taking our OdoNet for example. For our OdoNet, estimating only the forward speed of the vehicle rather than three-dimensional speed can also reduce the complexity of the DNN model, and thus improve the robustness and practicality.

III. PROPOSED METHOD

As an alternative to the wheeled odometer, OdoNet is a deep pseudo-odometer model based on one-dimensional (1D) CNN, which utilizes a window of IMU samples to estimate the vehicle speed. This section will describe the details of the proposed method, which includes the deep learning architecture of the OdoNet, the indispensable data-cleaning procedure, and the framework of the integrated navigation system enhanced by the OdoNet.

*A. OdoNet Architecture*

According to our literature review, various basic models have been applied for IMU-based DNN, including LSTM [6, 7, 9, 16, 17], Bi-LSTM [14, 16], CNN [5, 10, 15, 16], and even hybrid model [12, 13, 18]. RNN based networks, including LSTM and Bi-LSTM, are typically used for time-series data processing, e.g. time-series of IMU data. Although RNN-based networks might have better precision, their convergence speed is extremely low during model training, which greatly increases the verification cycle of the model. CNN has been proved to play an irreplaceable role in feature extraction, and it is widely used in computer vision analysis and even time-series data processing. According to our experiments in vehicle speed estimation, CNN outperforms RNN in both precision and efficiency running in a GPU, which will be proved in IV.B. Hence, we pick the 1D CNN model, and the network architecture of the OdoNet is shown in Fig. 2. As for the size of the window, many previous works [12, 13, 17] have evaluated its effects on their networks. If the window size is too small, it is hard to extract enough features to regress speed, while it might cause a long time delay if the window size is too large. For our OdoNet, experiment results demonstrate that one second is the most appropriate window size. Whilst, the output rate of IMU in our self-developed GNSS/INS integrated system is 50Hz, so the input data is treated as a 1D vector with a length of 50 and a depth of 6.

As shown in Fig. 2, the OdoNet consists of four 1D convolution layers, with max-pooling layers for sub-sampling. After flattening depth, three fully connected layers are employed and output the regressed speed. The activation functions between two layers are ReLU [32] units, except the pooling layers. Moreover, two dropout layers with probability of 0.5 are used before the first two fully connected layers, to avoid over-fitting during training. The kernel size, the depth of the convolution layer, and the units of the fully connected layer can be found in Fig. 2. Hence, the input and output of the OdoNet can be described as follows:

$$(\boldsymbol{w}_{iv}^v, \boldsymbol{f}^v)_{50*6} \xrightarrow{f_\theta} s \cdot \tilde{v}_{odo} \quad (1)$$

where $\boldsymbol{w}_{iv}^v$ and $\boldsymbol{f}^v$ are compensated IMU measurements as the input; $f_\theta$ is the OdoNet itself; $s$ is the scale factor of the output, and $\tilde{v}_{odo}$ is the regressed speed. For typical land vehicle applications, the maximum speed of the vehicle can only reach 30 m/s in an urban environment. Hence, we use a fixed scale factor of $s = 1/30$ to normalize the speed, which can accelerate the convergence of the model during training. To solve the optimal parameter $\theta^*$ inside the OdoNet, we can minimize the loss function $l$ on the training dataset as follows:

$$\theta^* = \arg\min l(f_\theta(\boldsymbol{w}_{iv}^v, \boldsymbol{f}^v), s \cdot v_{odo}) \quad (2)$$

The loss function is defined as the mean squared error (MSE) between the truth $s \cdot v_{odo}$ and the regressed value $s \cdot \tilde{v}_{odo}$ as follows:

$$l = \frac{1}{N} \sum (s \cdot v_{odo} - s \cdot \tilde{v}_{odo})^2 \quad (3)$$

where $N$ is the size of the IMU samples.

Since ReLU activation function [32] is applied in the network, which means that the output of the network is always greater than or equal to zero, this characteristic provides an opportunity for zero-velocity detection. In other words, the regressed speed is zero or close to zero during stationary states. Consequently, we can simply apply a fixed threshold detector to determine zero-velocity states. Mathematically, we can formalize the detection problem as a binary hypothesis testing problem, where the detector can choose between the following two hypotheses [1]:

$$H_0: \text{IMU is moving}$$
$$H_1: \text{IMU is stationary}$$
(4)

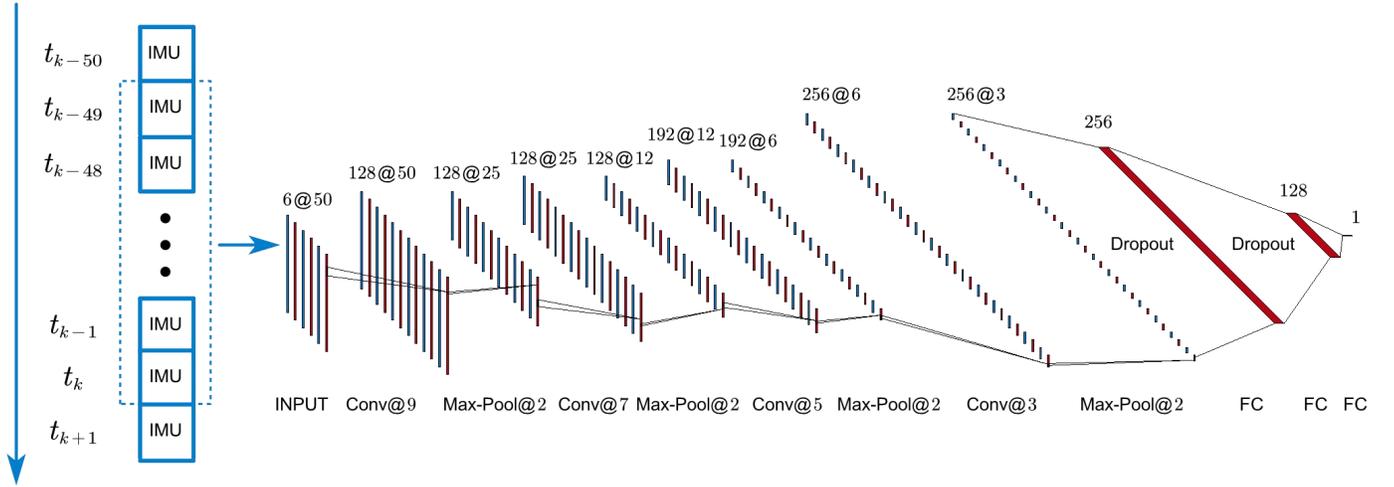

Fig. 2. Proposed OdoNet network architecture. The INPUT, Conv, Max-Pool and FC represent input layer, convolution layer, max pooling layer, and fully connected layer, respectively. The name of each layer is shown below the graph, with the kernel size for convolution and max pooling layers. The numbers above the network are the depth@length for convolution and max pooling layers, or the units for the fully connected layers. The schematic of the DNN is generated by [29].

According to the experience of engineering, the threshold of the zero-velocity detector is set to 0.1 m/s. Hence, the judgment for these two hypotheses can be written as follows:

$$H_0: \tilde{v}_{odo} \geq 0.1 \ m/s$$
$$H_1: \tilde{v}_{odo} < 0.1 \ m/s$$
(5)

It is generally known that the robustness and precision of a DNN are affected by various factors. The possible factors include the IMU biases, the mounting angles, the IMU individuality, the vehicle loads, and the road conditions. In section IV. D, experiment results will be presented to demonstrate the robustness and the generalization capability of the OdoNet.

*B. Data Cleaning*

According to our experiments, we find that the IMU biases and the mounting angles have strong impacts on the precision of the OdoNet. To mitigate or eliminate their effects, we compensate them precisely in this data-cleaning procedure, which has been proven to significantly improve the robustness and the precision of the OdoNet.

Low-cost MEMS IMU has extremely low stability, with poor temperature characteristics, which leads to huge and rapidly changing biases [25]. As a matter of experience, taking ICM20602 for example, the biases of the gyroscope typically range from -2 °/s to 2 °/s, and the biases of the accelerometer typically range from -0.2 m/s$^2$ to 0.2 m/s$^2$. Moreover, each IMU has different characteristics of biases, which might seriously affect the robustness and precision of the OdoNet. Improving the complexity of the model or using a larger dataset might reduce the impact of the IMU biases, but it involves heavy dataset collecting work and more computational resources. Fortunately, the GNSS/INS integrated system can effectively and accurately estimate the IMU biases, as in (6). With estimated gyroscope biases $\hat{\boldsymbol{b}}_g$ and accelerometer biases $\hat{\boldsymbol{b}}_a$, we can compensate gyroscope measurement $\tilde{\boldsymbol{w}}_{ib}^b$ and accelerometer measurement $\tilde{\boldsymbol{f}}^b$ in advance. $(\cdot)^b$ represents the vector in the IMU body frame (b-frame).

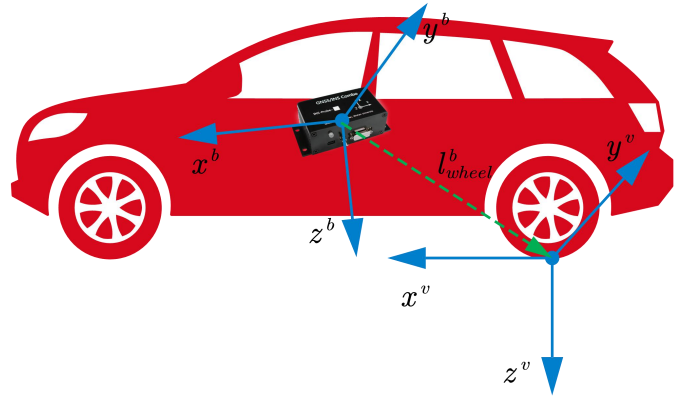

Fig. 3. Definitions of the vehicle frame (v-frame) and the IMU body frame (b-frame), and illustration of the IMU mounting angles and the lever-arms with respect to the v-frame.

As we all know, the calculated speed by INS in vehicle frame (v-frame) is the function of the lever-arms $\boldsymbol{l}_{wheel}^b$ and the IMU mounting angles $\mathbf{C}_b^v$, as can be seen in (10). Hence, the regressed speed estimated by the OdoNet should be in v-frame exactly, rather than in b-frame. In addition, for most vehicle navigation applications, the navigation devices (including the IMU) are usually mounted on the given spot of the vehicle, which means that the lever-arms are almost unchanged. However, the mounting angles of the IMU with respect to the vehicle might be randomly set and unknown, which will change the IMU signal significantly and ruin the OdoNet. Consequently, the mounting angles must also be compensated in preprocessing. The estimating algorithm of the mounting angles is well described in [26]. However, the roll mounting angle cannot be estimated due to a lack of observability [26], so the OdoNet must be able to adapt this factor. Estimated pitch mounting angle $\Delta\theta$ and heading mounting angle $\Delta\psi$ can be converted to direction cosine matrix $\mathbf{C}_b^v$. Then, the raw IMU measurements can be converted from b-frame to v-frame.

To sum up, in order to eliminate the impact of the unique IMU biases and the mounting angles, we compensate them in advance. More specifically, the IMU biases are derived from the EKF (6) in real-time and the mounting angles are estimated in the post-processing. Certainly, there are still residual errors in the compensated data, which means that OdoNet must be able to adapt uncompensated data with residual biases and mounting angles. In section IV. D, we will show evidence about the importance of data-cleaning procedure, and illustrate how these factors affect the robustness and precision of the OdoNet.

### C. Integrated Navigation System Framework

The proposed integrated navigation system for vehicle application is depicted in Fig. 1. In a conventional GNSS/INS integrated navigation system, raw IMU data is processed by the INS mechanization algorithm [27-28], and an EKF is applied to fuse GNSS and INS, combining with non-holonomic constraint or other constraints or aiding sources. However, without odometer aiding, it might still lead to large positioning drifts in GNSS-denied urban environments, especially in the forward direction of the vehicle. Based on the proposed OdoNet, the pseudo-odometer with a zero-velocity detector can be employed in the integrated navigation system. As can be seen in Fig. 1, raw IMU samples are compensated by estimated IMU biases from the EKF and prior estimated mounting angles from post-processing. With compensated IMU samples, the OdoNet predicts the raw vehicle speed, which is then filtered by an FIR (Finite Impulse Response) low-pass filter. According to experiments, the cut-off frequency of the FIR filter is set to 0.1 Hz and its order is set to 64. The filtered vehicle speed is treated as a pseudo-odometer measurement and is also used to determine the zero-velocity states of the vehicle. Then, the pseudo-odometer speed update, the zero velocity update (ZUPT), and the zero angular rate update (ZARU) can be employed in EKF. With our proposed pseudo-odometer and zero-velocity detector, the GNSS/INS integrated navigation system can be constrained effectively, and thus can maintain continuous high-accuracy positioning in GNSS-denied environments.

The error state vector of the used extend Kalman filter is defined as follows:

$$\boldsymbol{x} = [\,\delta\boldsymbol{r}^n \quad \delta\boldsymbol{v}^n \quad \boldsymbol{\psi} \quad \delta\boldsymbol{b}_g \quad \delta\boldsymbol{b}_a\,]^T \tag{6}$$

where $\delta\boldsymbol{r}^n$ and $\delta\boldsymbol{v}^n$ can be written as follows:

$$\delta\boldsymbol{r}^n = [\,\delta r_N \quad \delta r_E \quad \delta r_D\,]^T \tag{7}$$

$$\delta\boldsymbol{v}^n = [\,\delta v_N^n \quad \delta v_E^n \quad \delta v_D^n\,]^T \tag{8}$$

$\delta\boldsymbol{r}^n$ represents the position error in navigation frame (n-frame) for the NED (North-East-Down) coordinate system; $\delta\boldsymbol{v}^n$ represents the velocity error; $\boldsymbol{\psi}$ denotes the attitude error using the psi-angle model [27]; $\delta\boldsymbol{b}_g$ and $\delta\boldsymbol{b}_a$ represent the gyroscope and accelerometer biases error, respectively. The state equation, the GNSS positioning observation equation, the implementation of the Kalman filter, and more detailed information can refer to [27-28].

NHC is a kind of virtual velocity observation in the v-frame, which means that there is no lateral and vertical speed for the land vehicle. The odometer can provide forward speed aiding, and thus NHC and odometer can be combined to achieve three-dimensional velocity aiding in v-frame. The measured velocity in v-frame can be written as follows:

$$\tilde{\boldsymbol{v}}_{wheel}^v = [\,\tilde{v}_{odo}^v \quad 0 \quad 0\,]^T + \boldsymbol{e}_v \tag{9}$$

The calculated velocity in v-frame by INS can be expressed as follows:

$$\hat{\boldsymbol{v}}_{wheel}^v = \mathbf{C}_b^v \hat{\mathbf{C}}_n^b \hat{\boldsymbol{v}}_{IMU}^n + \mathbf{C}_b^v (\hat{\boldsymbol{w}}_{nb}^b \times) \boldsymbol{l}_{wheel}^b \tag{10}$$

where $(\hat{a})$ denotes calculated parameters by INS; $\tilde{v}_{odo}^v$ denotes the speed of the odometer; $\boldsymbol{e}_v$ denotes the observation noise; $\mathbf{C}_b^v$ represents the IMU mounting angles with respect to the vehicle frame; $\boldsymbol{l}_{wheel}^b$ is the lever-arms vector between b-frame and v-frame in b-frame. Typically, the measurement center of velocity observation in the v-frame is defined at the point in the rear wheel, where the wheeled odometer is installed, and it is tangent to the ground, as depicted in Fig. 3.

When the stationary state is determined by the zero-velocity detector, ZUPT and ZARU can be applied to EKF. Zero velocity means that the IMU or the vehicle is stationary, and it can be written as:

$$\tilde{\boldsymbol{v}}^n = \boldsymbol{e}_v \tag{11}$$

By using ZUPT during the stationary state, the drift of velocity and position can be constrained effectively, except the heading angle, and thus ZARU must be used simultaneously. Zero angular rates can be expressed as:

$$\tilde{\boldsymbol{w}}_{ib}^b = \boldsymbol{e}_w \tag{12}$$

where $i$ denotes the inertial frame (i-frame), and $\boldsymbol{e}_w$ represents the observation noise.

According to our dedicated experiments, the standard deviation values are set to at 0.1 m/s for both NHC and ZUPT and 0.1 °/s for ZARU. As for the odometer, we should consider its precision. Specifically, the standard deviation value of the real wheeled odometer is set according to its resolution, while the value for the pseudo-odometer is derived from its evaluation metrics. However, the basic hypothesis of NHC is not satisfied occasionally in some extreme cases, like slippage. The OdoNet might fail to estimate and present wrong speed with large error, and the zero-velocity detector might also give wrong judgment. Consequently, a fault detection algorithm must be applied to eliminate gross error in these observations. We use innovation [28] in the extended Kalman filter to detect and reject faults. For an EKF, the normalized innovations are defined as:

$$y_{k,j}^- = \frac{\delta z_{k,j}^-}{\sqrt{\mathbf{C}_{k,j,j}^-}} \tag{13}$$

where $\delta z_{k,j}^-$ is the measurement innovation, $C_{k,j,j}^-$ is the covariance of the innovation. For a one-dimensional observation, if the degree of confidence is set to 95%. The observations with $y_{k,j}^- > 3.84$ will be judged as outliers or gross errors, and will not be employed in the EKF.

## IV. EXPERIMENTS AND RESULTS

This section examines the proposed OdoNet and the integrated navigation system enhanced by it. A comprehensive vehicle dataset was collected to train and evaluate the OdoNet. After preprocessing the dataset, the model was trained. Then, the metrics of the regressed speed and zero-velocity detection on the testing dataset were evaluated. Next, dedicated experiments were carried out to evaluate the possible factors that might affect the robustness of OdoNet, so as to verify the generalization capability of the OdoNet. Finally, we compared the performance of the integrated navigation system on three typical testing sequences using three different aiding sources, to evaluate the contribution of the proposed pseudo-odometer. Experiments mentioned below, except the dataset collection, were all done offline.

### A. Dataset Collection

As shown in Fig. 4, the whole sensors setup contains 9 INS-Probe (our self-developed GNSS/INS integrated navigation system) modules, a high-precision wheeled odometer, and a navigation-grade POS (Position Orientation System) POS-A15. INS-Probe modules are responsible for collecting multi-sensors data, which includes GNSS RTK positioning from low-cost GNSS module (ZED-F9P), raw IMU data from MEMS IMU chip (ICM20602), and wheeled odometer (DFS60E). These sensors are precisely synchronized through hardware triggering. The output rate of the IMU is 50 Hz, and the wheeled odometer is synchronously sampled with IMU. The wheeled odometer is an optic encoder, DFS60E from SICK with 2048 lines, which will be used as a benchmark for the proposed OdoNet. POS-A15 is a reference system to evaluate the integrated navigation system, which can provide extremely high-precision ground truth.

The whole dataset consists of 14 test sequences, covering various areas in Wuhan city. As depicted in Fig. 5, our dataset contains most of the typical scenes, including freeway, university campus, residential districts, industrial park, and other different testing scenes in urban environments. 9 INS-Probe modules were used, and thus the data of 9 individual IMUs were collected, which greatly enriches the diversity of the dataset. We split the dataset into two parts, one for training and one for testing. The training dataset contains 10 different sequences per INS-Probe module from 7 modules, that is 70 sequences in total. The testing dataset is collected from the rest two INS-Probe modules, each with 4 sequences, that is 8 sequences in total.

The labels for model training were generated from the real wheeled odometer. Statistical analysis was carried out on the prepared dataset, as showed in Fig. 6. The distributions of the dataset show that the speeds of most samples are lower than 15 m/s. There are a total of 468786 samples for training, and 56744 samples for testing. The collected dataset contained various road conditions, where the vehicle traveled at different speeds in different scenes. Specifically, the freeway is a high-speed scene, and the university campus and industrial park are low-speed scenes and median-speed scenes respectively.

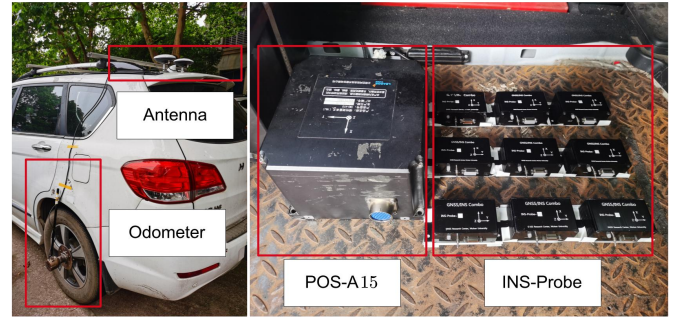

Fig. 4. Photograph of the test vehicle and system setup inside and outside the vehicle.

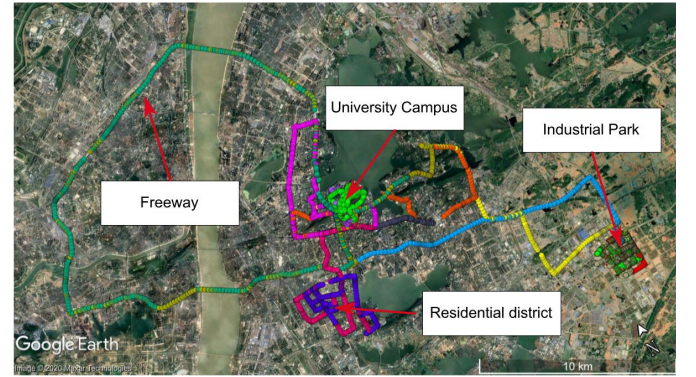

Fig. 5. The trajectory of the collecting dataset. Different colors represent different sequences. The map is generated by Google Earth.

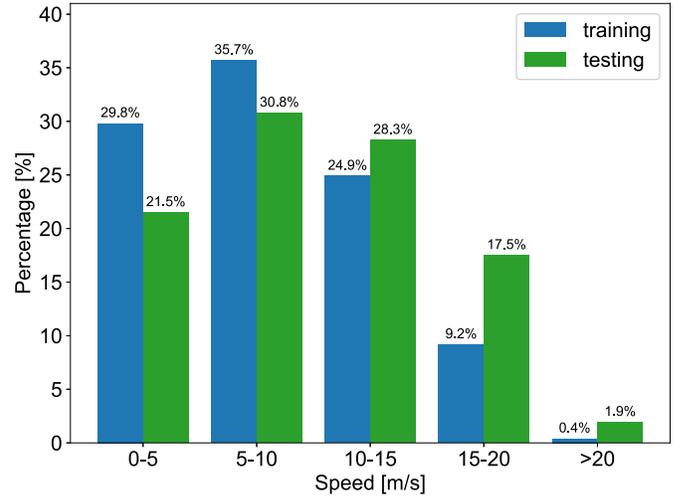

Fig. 6. The distributions of the speed on training and testing dataset.

### B. Model Training

Our model was implemented based on the public available Keras framework with TensorFlow. The training was done on an NVIDIA TITAN Xp GPU. The training dataset was shuffled randomly, and 20% of them were used as validation dataset during the training. The hyperparameters in model training were derived by applying grid search to evaluate validation results. As a result, the Adam optimizer [30] was used with a learning rate of 0.00005, and a big batch size of 1024 samples was used for better generalization capability and higher precision.

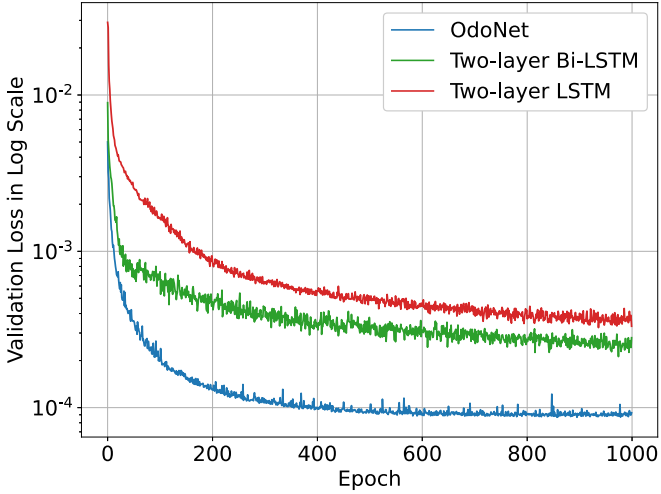

Fig. 7. The validation loss of the OdoNet and other two networks during the training. The loss is in log scale. The Bi-LSTM based network and LSTM based network have the same structures with OdoNet, except the four-layer 1D CNN network. Each layer of the Bi-LSTM has 64 nodes, and each layer of the LSTM has 128 nodes.

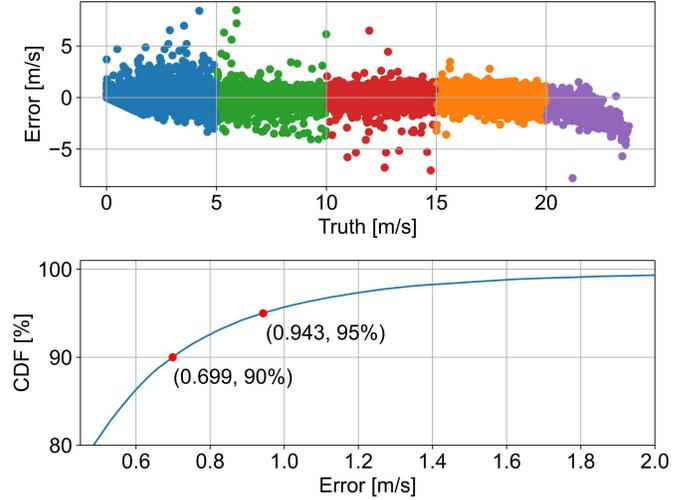

Fig. 8. The distributions of the error of regressed speed and the cumulative distribution function on the testing dataset.

TABLE I
THE ERROR OF REGRESSED SPEED ON TESTING DATASET

| Velocity [m/s] | 0-5 | 5-10 | 10-15 | 15-20 | 20-25 | Overall |
|---|---|---|---|---|---|---|
| MAE [m/s] | 0.300 | 0.253 | 0.294 | 0.397 | 1.070 | **0.315** |
| RMSE [m/s] | 0.534 | 0.391 | 0.419 | 0.520 | 1.337 | **0.490** |

We trained OdoNet for 1000 epochs, costing 7 seconds for each epoch. We also trained a two-layer LSTM network and a two-layer Bi-LSTM network. The LSTM based network and Bi-LSTM based network have the same structures as the OdoNet, except the four-layer 1D CNN network. Each layer of the LSTM has 128 nodes, and each layer of the Bi-LSTM has 64 nodes. As depicted in Fig. 7, the validation loss of the OdoNet converged rapidly with satisfying precision. In contrast, the two-layer LSTM network and the two-layer Bi-LSTM have lower convergence speed and precision, even if we have used a larger learning rate of 0.001. In addition, the LSTM and Bi-LSTM consumed much more computational resources, which led to longer training and predicting time. More specifically, the two-layer LSTM network took about 4 times as much time as the OdoNet for each epoch during training, and the value for the two-layer Bi-LSTM was about 8 times. All the above experiment results demonstrate that our CNN-based OdoNet has better performance with higher efficiency, compared to the RNN-based networks.

## C. Regressed Speed Evaluation

*1) Raw Regressed Speed on Testing Dataset:* With well-trained OdoNet, we evaluated the regressed speed on the testing dataset. The error of the regressed speed are shown in Fig. 8, where the first subplot depicts the distribution of the regressed errors, and the second subplot depicts the cumulative distribution function of the speed errors. Most of the speed errors are satisfying, except for a small number of gross errors. In the testing dataset, 90% of the regressed speed error are less than 0.699 m/s, and 95% of them are less than 0.943 m/s.

More specifically, the Mean Absolute Error (MAE) and Root Mean Square Error (RMSE) of the regressed speed are listed in Table I. The MAE and RMSE on the overall testing dataset are 0.315 m/s and 0.490 m/s respectively in a range of 0-25 m/s. However, we cannot ignore the fact that the precision of the OdoNet degrades seriously when the speed is higher than 20 m/s, which can be seen in both Fig. 8 and Table I. According to our analysis, lacking enough samples of high speed is the main reason. As depicted in Fig. 6, the samples of speeds higher than 20 m/s in total are less than 2%. Adding more high-speed samples should be able to solve this issue. The scenes where the vehicle can travel at a high speed of more than 20 m/s are generally wide roads away from tall buildings, e.g. highway, open-sky environments specifically. In general, GNSS can attain high-accuracy RTK positioning in such open-sky areas, and thus these low-precision regressed speeds should have a limited impact on the final results of the integrated navigation system or even be eliminated by the gross error detection algorithm (13).

*2) Regressed Speed Sequences on Testing Sequences:* For each 50 consecutive IMU samples, the OdoNet can predict a corresponding vehicle speed, and thus we can attain speed with an update rate of 50 Hz. The speed sequence is then filtered by the FIR low-pass filter. The testing dataset consists of two INS-Probe modules with 4 different sequences, including three sequences in different speed scenes and one sequence for robustness testing. These three scenes are mentioned in IV. A, and the details of the scenes can refer to IV. E. The RMSE of the raw regressed speed and filtered speed are shown in Table II, which shows great consistency between these two individual IMU. However, the RMSE of the regressed speed in the high-speed scene is much larger than the other two scenes, which corresponds to the previous results in Table I.

TABLE II
RMSE OF THE SPEED SEQUENCES IN THREE SCENES

| Scene | M1 | | M2 | |
|---|---|---|---|---|
| | Raw [m/s] | Filtered [m/s] | Raw [m/s] | Filtered [m/s] |
| Low Speed | 0.457 | 0.339 | 0.449 | 0.329 |
| Median Speed | 0.381 | 0.240 | 0.374 | 0.249 |
| High Speed | 0.618 | 0.486 | 0.594 | 0.449 |

M1 and M2 represents two different INS-Probe modules in the testing dataset. Raw denotes the raw regressed speed, and filtered denotes the filtered speed.

TABLE III
RESULTS OF ZERO VELOCITY DETECTION USING REGRESSED SPEED

| Metrics | Value | Metrics | Value |
|---|---|---|---|
| True Positive | 3636 | False Positive | 178 |
| True Negative | 52895 | False Negative | 35 |
| Precision | 95.33% | Recall | 99.05% |

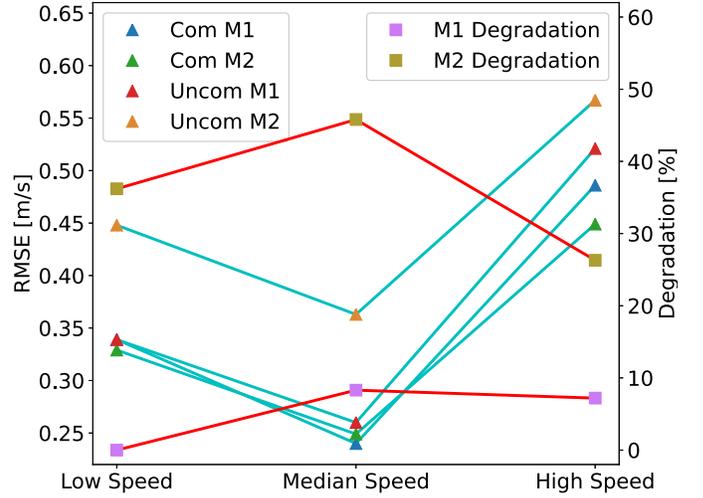

Fig. 9. The RMSE of the filtered speed using compensated model and uncompensated model. Com denotes that all the IMU sequences are well compensated in training and testing, while Uncom denotes no compensation. The degradation rate is defined as (Uncom-Com)/Com. The uncompensation is only for the IMU biases, and the mounting angles are compensated for all sequences in this test. M1 and M2 represents the two different INS-Probe modules in testing dataset.

*3) Zero Velocity Detection on Testing Dataset:* We also evaluated the performance of the zero-velocity detector. We used (7) to generate the ground truth of the zero-velocity states from the real wheeled odometer. The evaluation metrics for binary classification on the testing dataset are shown in Table III. The precision and recall are 95.33% and 99.05% respectively, which illustrates that the false-alarm rate is slightly larger than the missed-detection rate. The results are reasonable because the regressed speeds are mixed with gross errors. ZUPT and ZARU are extremely important in some scenes because they can improve the precision and stability of the integrated system effectively. Moreover, a false alarm of zero velocity might be eliminated by the fault-detection model in (13). In other words, we need higher recall to a certain extend, and thus the results of the zero-velocity detector are acceptable.

*D. Robustness Evaluation*

In the previous section, we evaluate the regressed speed metrics, and the results illustrate that the raw regressed speed on the testing dataset is less than 0.5 m/s in a wide dynamic range of 0-25 m/s. However, the most common challenges for a deep-learning network are the robustness and the generalization capability, which determine its usability in more general scenes. In this section, we evaluate some common factors that might affect the robustness of OdoNet, including the compensation of the IMU biases and the mounting angles, the vehicle loads, and the road conditions. Previous experiments have demonstrated that the IMU individuality has little impact on the robustness and precision of the OdoNet, so it will not be evaluated repeatedly. The M1 and M2 mentioned below represent the two INS-Probe modules in the testing dataset.

*1) Impact of Compensation:* In the data-cleaning procedure, the IMU biases and the mounting angles are precisely compensated, where the IMU biases are derived from EKF in real-time and the mounting angles are estimated in post-processing. It is reasonable that IMU biases have a great impact on the robustness of the model. What we need is the forward speed of the vehicle, and thus the mounting angles of the IMU to the vehicle might also affect the generalization capability of the OdoNet.

The biases of a single IMU are unique and different from each other, so we trained and tested the OdoNet with raw IMU samples (without compensating for the biases) to evaluate their impact. We compared the RMSE of the filtered regressed speed of two different models in three testing scenes, as depicted in Fig. 9. The impacts on M1 and M2 are completely different. The uncompensated model has little impact on the precision of the M1, and the RMSE increase by less than 10%. However, the precision of M2 on uncompensated models degrades as much as 45%. A possible explanation might be the insufficient IMU individuality in the training dataset, which is uncontrollable due to the uncertainty of the IMU biases. In contrast, it is convenient to attain IMU biases from the integrated navigation system online. Hence, training with biases calibrated IMU data makes the OdoNet more robust and more precise.

The mounting angles of the IMU in our dataset are small, and each axis is less than 2° typically (according to the estimated results), mainly because the IMU is mounted aligning to the vehicle. Consequently, we manually added extra mounting angles to the IMU samples and evaluated the RMSE of the regressed speed. The data in the median-speed scene is used in this test. As we can see in Fig. 10, the pitch and heading mounting angles have a completely different impact on the precision of the regressed speed. The impact of the pitch mounting angle is much larger than the heading mounting angle. If 0.4 m/s is the acceptable precision, the tolerable pitch mounting angle is about 2°, while the tolerable heading mounting angle is about 12°. It can be explained that the acceleration is directly related to the vehicle speed, and the pitch mounting angle might lead to the change of the gravity

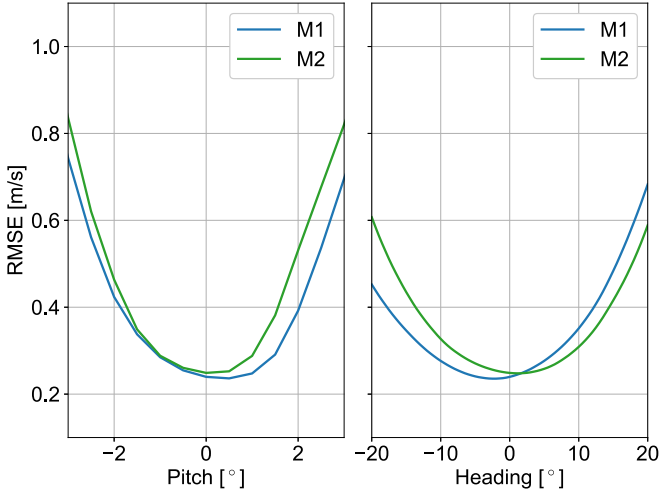

Fig. 10. The RMSE of the filtered speed with the change of mounting angles in the median-speed scene. We evaluated pitch and heading mounting angles separately.

TABLE IV
RMSE OF THE SPEED SEQUENCES WITH DIFFERENT VEHICLE LOADS

| Loads | M1 | | M2 | |
| --- | --- | --- | --- | --- |
| | Raw [m/s] | Filtered [m/s] | Raw [m/s] | Filtered [m/s] |
| Two People | 0.389 | 0.259 | 0.441 | 0.296 |
| Three People | 0.423 | 0.304 | 0.443 | 0.307 |
| Four People | 0.410 | 0.260 | 0.472 | 0.311 |

The number of people includes the driver himself. M1 and M2 represent the two INS-Probe modules in the testing dataset.

component on different axes of the accelerometer. The error of the estimated pitch mounting angle is typically within 0.005° [26] for low cost MEMS IMU, while the error of the estimated heading mounting angle is much larger. As long as the accuracy of the estimated mounting angles is within 2°, the OdoNet is capable of adapting these factors easily and maintains reasonable accuracy of the regressed speed.

*2) Impact of Vehicle Loads:* Different loads might affect the dynamic characteristics of the vehicle in some cases, and thus the vehicle loads might affect the precision of OdoNet to some extend. A special experiment was carried out to evaluate the impact of the vehicle loads. We drove the vehicle on the same road on a round trip with different loads (i.e. with a different number of passengers), and the length of the road was about 6 km. Three tests were conducted, and the number of passengers was two, three, and four, respectively. The RMSE of the regressed speed in each test is displayed in Table IV. No notable difference is found between these three tests, not only from the raw-speed error but also from the filtered-speed error, which illustrates that the impact of the vehicle loads is negligible.

*3) Impact of Road Conditions:* It is obvious that different road conditions might affect the motion of the vehicle, and thus might affect the precision of the OdoNet. Even though our dataset has covered various roads in Wuhan city, we should

TABLE V
RMSE OF THE SPEED SEQUENCES IN NEW SCENES

| Sequences | Raw [m/s] | Filtered [m/s] |
| --- | --- | --- |
| Seq. 1 | 0.472 | 0.336 |
| Seq. 2 | 0.443 | 0.333 |
| Seq. 3 | 0.377 | 0.277 |

Seq. n denotes different testing sequences.

evaluate the OdoNet with data collected in totally new scenes. Three new IMU sequences were used in this test. These testing sequences were collected in a totally strange scene by a new IMU, specifically, another industrial district in Wuhan city. As shown in Table V, the RMSE of the regressed vehicle speed on these three sequences are very close, mainly because of the same testing scene. Moreover, the results show nearly the same accuracy compared to previous results in Table II, which demonstrates the superior generalization capability of the OdoNet to different road conditions. Moreover, the experiment results illustrate that the OdoNet is robust to different IMU individuality once again.

*E. Navigation Results*

In this part, we will present the results of the GNSS/INS integrated navigation using three different aiding modes, including the NHC-only, the pseudo-odometer aiding, and the wheeled odometer aiding. By comparing the navigation results of these three processing modes, we can judge the precision of our proposed method. When applying odometer aiding, NHC is always applied simultaneously. The difference between the NHC-only mode and the odometer mode is in the longitudinal direction of the vehicle instead of the literal, and thus we can use horizontal error to represent the positioning error. The M1 and M2 mentioned below represent the two INS-Probe modules in the testing dataset.

In addition, ZUPT and ZARU are used only when the odometer is used, because the zero-velocity state is determined by the odometer speed. The standard deviation of pseudo-odometer observation is set to 0.3 m/s, according to the RMSE of the filtered speed in Table II, while 0.1 m/s for real wheeled odometer according to its precision.

The GNSS results used in experiments were derived from ZED-F9P working in RTK mode. In GNSS complex environments, only those GNSS positioning results with a standard deviation less than 1 m were employed in the integrated navigation system.

*1) Low-speed Scene:* Due to narrow roads and complex traffic conditions, the vehicle traveled very slowly around the university campus. In this low-speed scene, there was intensive coverage of tall trees along the roads, which led to long GNSS outages occasionally. The longest three GNSS outages lasted 36, 56, and 131 seconds, respectively. Fig. 11 shows the filtered regressed speeds of M1 and their error. As we can see, the highest speed is only about 10 m/s, without stationary states during the travel. Most of the filtered speeds are precise, except for some gross error exceeding 1 m/s.

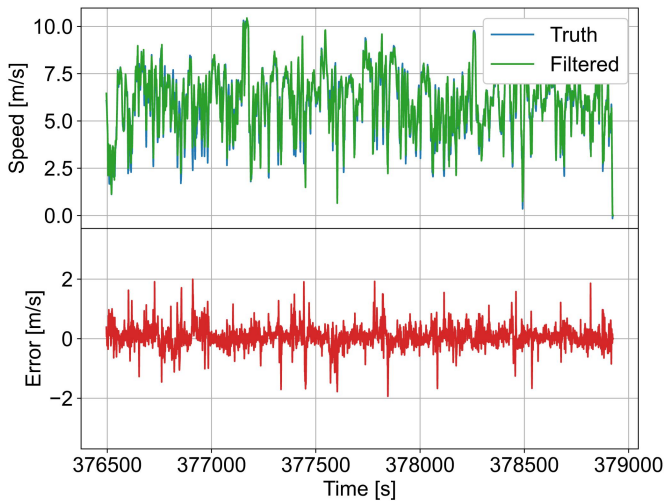

Fig. 11. The filtered regressed speed of the testing M1 and its error in the low-speed testing scene.

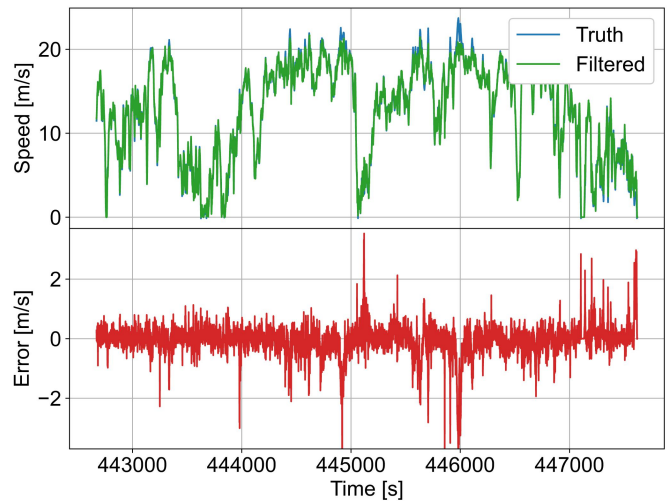

Fig. 13. The filtered regressed speed of the testing M1 and its error in the high-speed testing scene.

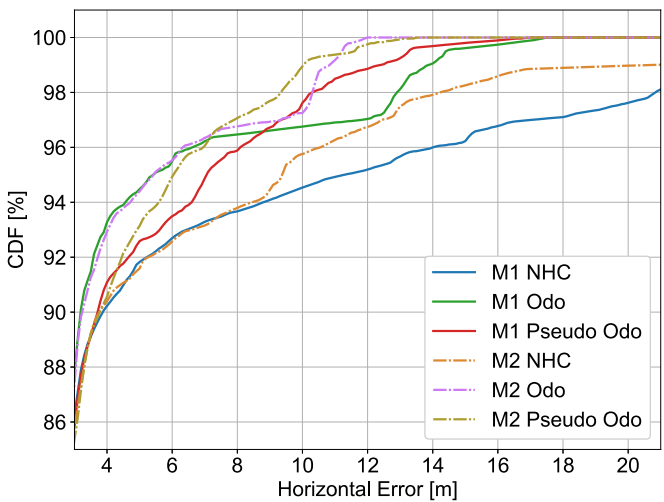

Fig. 12. The cumulative distribution function of horizontal error in the low-speed testing scene. The M1 and M2 represents two different INS-Probe modules in testing dataset. The Odo denotes the wheeled odometer aiding. The Pseudo Odo denotes the pseudo-odometer aiding from OdoNet.

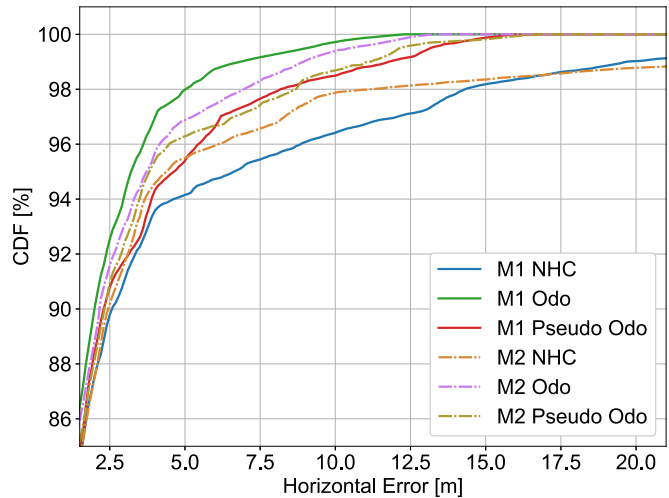

Fig. 14. The cumulative distribution function of horizontal error in the high-speed testing scene.

Fig. 12 shows the cumulative distribution function (CDF) of the two testing modules with three processing modes. Due to higher precision, the wheeled odometer improves the precision of navigation significantly. The pseudo-odometer also shows great improvement compared to the NHC-only mode. The precision of the pseudo-odometer is extremely close to the real wheeled odometer in the low-speed scene. However, as we can see, the pseudo-odometer only takes effects during long GNSS outages, which means that the integrated navigation system might have a large drift without pseudo-odometer aiding. The most important thing is to reduce large drift in vehicle navigation, and the pseudo-odometer and the wheeled odometer can play the same role on this issue.

*2) High-speed Scene:* The high-speed scene is the Second Ring Road of Wuhan city, without any traffic light. There are four long tunnels during the travels, with some other GNSS-challenged environments. The longest three GNSS outages lasted 73, 80, and 156 seconds, respectively. The filtered regressed speed and their error are shown in Fig. 13. The vehicle traveled at a very high speed, except when traffic jams occurred. As mentioned in IV. C, the OdoNet shows poor precision when the speed exceeds 20 m/s; the regressed speed contains a large error at 446000s in Fig. 13, for example. Besides, the RMSE of both the raw and the filtered speed in the high-speed scene are larger than that of the other two scenes, as can be seen in Table II.

Fig. 14 shows the CDF of the horizontal error in the high-speed scene. The pseudo-odometer still outperforms the NHC-only mode, but cannot achieve the precision as the same as the real wheeled odometer. However, The pseudo-odometer shows outstanding performance in reducing large drifts. As depicted in Fig. 14, 100% of the error is less than 17 m after applying the pseudo-odometer aiding or the wheeled odometer aiding. However, if only the NHC is used, the maximum error is far larger than 20 m when the vehicle travels through the long tunnels, where GNSS tends to have long outages. All in all, the pseudo-odometer and the wheeled odometer almost make the same contribution to reducing large positioning drifts.

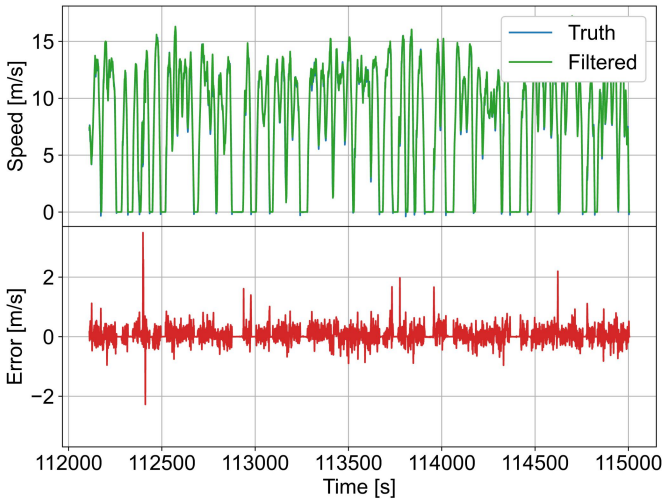

Fig. 15. The filtered regressed speed of the testing M1 and its error in the median-speed testing scene.

TABLE VI
THE RMS OF HORIZONTAL ERROR DURING MIMIC GNSS OUTAGES

| Module | Aiding | Seq. 1 [m] | Seq. 2 [m] | Seq. 3 [m] | RMS [m] |
|---|---|---|---|---|---|
| M1 | NHC | 15.624 | 8.974 | 10.669 | 12.090 |
| M1 | Pseudo Odo | **5.151** | **3.024** | **3.713** | **4.061** |
| M1 | Odo | 4.018 | 2.273 | 2.780 | 3.111 |
| M2 | NHC | 11.351 | 8.572 | 16.364 | 12.518 |
| M2 | Pseudo Odo | **3.567** | **3.894** | **4.005** | **3.827** |
| M2 | Odo | 3.367 | 2.752 | 3.359 | 3.172 |

Seq. n denotes three different outage sequences, which means that outages have different start time. Seq. 1 and Seq. 2 each has 13 outages, and Seq. 3 has 12 outages. The length of each GNSS outage is 60 seconds. The last column represents the RMS of the results of the three sequences.

*3) Median-speed Scene:* This scene is located in an industrial district of Wuhan city, with no tall buildings and few other vehicles. We drove the vehicle back and forth on these roads, and we often had to stop at the crossroads waiting for the traffic lights. As depicted in Fig. 15, the vehicle traveled steadily, without violent motions, either running around a fixed speed or staying stationary. The error of the filtered regressed speed is extremely small. The RMSE of the filtered speed is only 0.240 m/s and 0.249 m/s for the two testing modules respectively, which outperforms that in the other two scenes.

This scene is an open-sky environment, and thus GNSS can achieve centimeter-level RTK positioning. In this testing scene, we evaluated the performance of the integrated system by mimic GNSS outages [31]. We interrupted the GNSS positioning update in EKF for 60 seconds with a period of 180 seconds. During each outage, we picked up the maximum horizontal positioning error and then calculated the RMS of all the maximum errors in each outage. Three tests were conducted with a different start time of the outages. Table VI shows the experiment results of these two testing modules in three processing modes. As we can see, compared to the NHC-only mode, the pseudo-odometer significantly improves the precision of the positioning, and the RMSE are reduced by 66.4% and 69.4% for the two testing modules respectively. In fact, ZUPT and ZARU also make contributions to the improvement, which benefits from the reliable and effective zero-velocity detector using the pseudo-odometer speed.

The experiment results in the median-speed scene are statistically significant, and thus we can obtain quantitative conclusions. More specifically, the RMSE of the two testing modules in the NHC-only mode, the pseudo-odometer mode, and the wheeled-odometer mode, are 12.31m, 3.95m, and 3.14m, respectively. Compared to the NHC-only mode, by using the pseudo-odometer, the horizontal position drift error is reduced by about 68%, while the percentage for the wheeled odometer is about 74%. Generally speaking, the proposed pseudo-odometer can act as an alternative to the real wheeled odometer.

## V. CONCLUSIONS AND DISCUSSIONS

In this paper, we propose OdoNet, a untethered CNN-based deep pseudo-odometer model for vehicle GNSS/INS integrated navigation system. Comprehensive experiment results demonstrate that the OdoNet is robust enough to adapt to various changing conditions, including the IMU individuality, the vehicle loads, and the road conditions, while the IMU biases and the mounting angles can be compensated in data-cleaning procedure to mitigate their impacts. Navigation results indicate that after using the pseudo-odometer, the horizontal position drift error (60-second GNSS outage) is reduced by around 68% (compared to the NHC-only mode), which is very close to that of the hardware wheeled odometer. In conclusion, the proposed pseudo-odometer can significantly improve the accuracy and the practicality of the integrated navigation system, and it is a reliable alternative to the hardware wheeled odometer.

As we all know, the hardware wheeled odometer may be invalid during slipping or skidding, while the pseudo-odometer can still work. Hence, it is a meaningful work to combined the wheeled odometer and the pseudo-odometer to achieve an accurate and reliable navigation. Furthermore, it's also worth to explore the essence of the OdoNet, and we are intend to find out what the OdoNet has learned from the IMU data.

ACKNOWLEDGEMENT

This research is funded by the National Key Research and Development Program of China (No. 2020YFB0505803), and the National Natural Science Foundation of China (No. 41974024). The authors would also like to thank Dr. Xin Feng in our group for proofreading the paper.

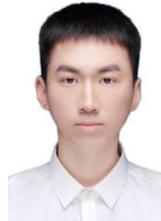

**Hailiang Tang** received the B.E. and M.E. degrees from Wuhan University, China, in 2017 and 2020, respectively. He is currently pursuing the Ph.D. degree in communication and information system with the GNSS Research Center, Wuhan University.

His current research interests include GNSS/INS integration technology, deep learning, visual SLAM, and autonomous robotics system.

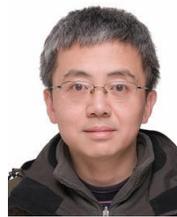

**Xiaoji Niu** received his bachelor's and Ph.D. degrees from the Department of Precision Instruments, Tsinghua University, in 1997 and 2002, respectively. He is currently a Professor with the GNSS Research Center, Wuhan University, China.

He did post-doctoral research with the University of Calgary and worked as a Senior Scientist in SiRF Technology Inc. He has published more than 90 academic papers and own 28 patents. He leads a multi-sensor navigation group focusing on GNSS/INS integration, low-cost navigation sensor fusion, and its new applications.

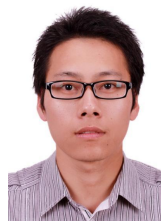

**Tisheng Zhang** is an associate professor in GNSS Research Center at Wuhan University, China. He holds a B.SC. and a Ph.D. in Communication and Information System from Wuhan University, Wuhan, China, in 2008 and 2013, respectively. From 2018 to 2019, he was a PostDoctor of the HongKong Polytechnic University.

His research interests focus on the fields of GNSS receiver and multi-sensor deep integration.

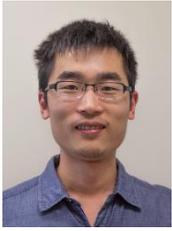

**You Li** is a Professor at the State Key Laboratory of Information Engineering in Surveying, Mapping and Remote Sensing (LIESMARS), Wuhan University, China. He received Ph.D. degrees from Wuhan University and University of Calgary in 2015 and 2016, respectively, and a BEng degree from China University of Geoscience (Beijing) in 2009. His research focuses on positioning and motion-tracking techniques and their uses in IoT devices, smartphones, robots, and cars. He has hosted/participated in five national research projects, and co-published over 70 academic papers, and has over 20 patents pending. He serves as an Associate Editor for the IEEE Sensors Journal, a committee member at the IAG unmanned navigation system and ISPRS mobile mapping working groups.

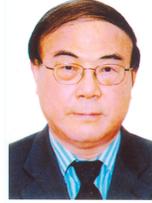

**Jingnan Liu**, member of Chinese Academy of Engineering, professor, Ph. D supervisor. He is an expert in geodesy and surveying engineering with the specialty of GNSS technology and applications. He has been engaged in the research of geodetic theories and applications, including national coordinate system establishment, GNSS technology and software development, as well as large project implementation. Over the past few decades, he has been engaged in the research of geodetic theories and applications. And so far he has published more than 150 academic papers and supervised more than 100 postgraduates.